\documentclass{bmvc2k}
\usepackage{times}
\usepackage{epsfig}
\usepackage{graphicx}
\usepackage{amsmath}
\usepackage{amssymb}
\usepackage{multirow}
\usepackage{subfigure}

\title{Forecasting Future Action Sequences with Neural Memory Networks}

\addauthor{Harshala Gammulle}{pranali.gammule@qut.edu.au}{1}
\addauthor{Simon Denman}{s.denman@qut.edu.au}{1}
\addauthor{Sridha Sridharan}{s.sridharan@qut.edu.au}{1}
\addauthor{Clinton Fookes}{c.fookes@qut.edu.au}{1}

\addinstitution{
Image and Video Research Laboratory (SAIVT)\\ 
Queensland University of Technology\\
Australia.
}

\runninghead{Gammulle et al.}{Forecasting Future Action Sequences with NMNs}

\begin{document}

\maketitle

\begin{abstract}
We propose a novel neural memory network based framework for future action sequence forecasting. This is a challenging task where we have to consider short-term, within sequence relationships as well as relationships in between sequences, to understand how sequences of actions evolve over time. To capture these relationships effectively, we introduce neural memory networks to our modelling scheme. We show the significance of using two input streams, the observed frames and the corresponding action labels, which provide different information cues for our prediction task. Furthermore, through the proposed method we effectively map the long-term relationships among individual input sequences through separate memory modules, which enables better fusion of the salient features. Our method outperforms the state-of-the-art approaches by a large margin on two publicly available datasets: Breakfast and 50 Salads. 
\end{abstract}

\vspace{-3mm}
\section{Introduction}
\vspace{-3mm}
 
\begin{figure}[htbp]
        \centering
       \subfigure[Action Anticipation]{\includegraphics[width=0.49\linewidth]{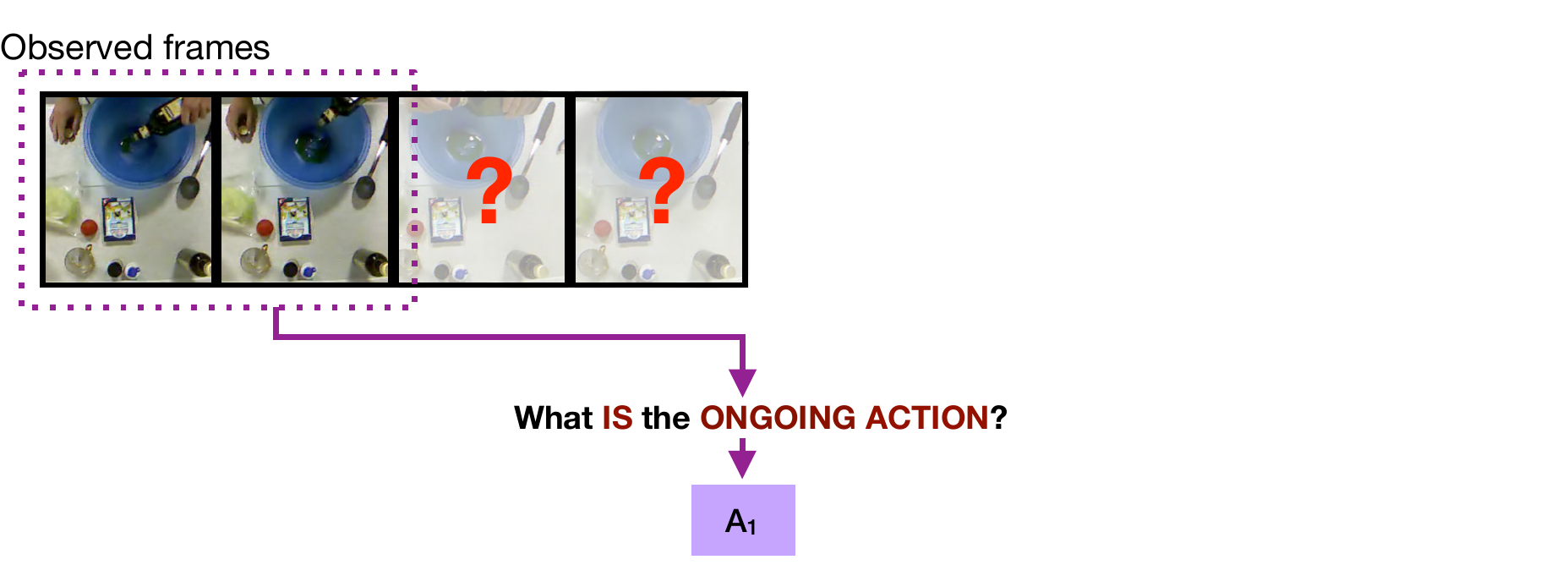}} 
	\subfigure[Future Action Sequence Prediction]{\includegraphics[width=0.49\linewidth]{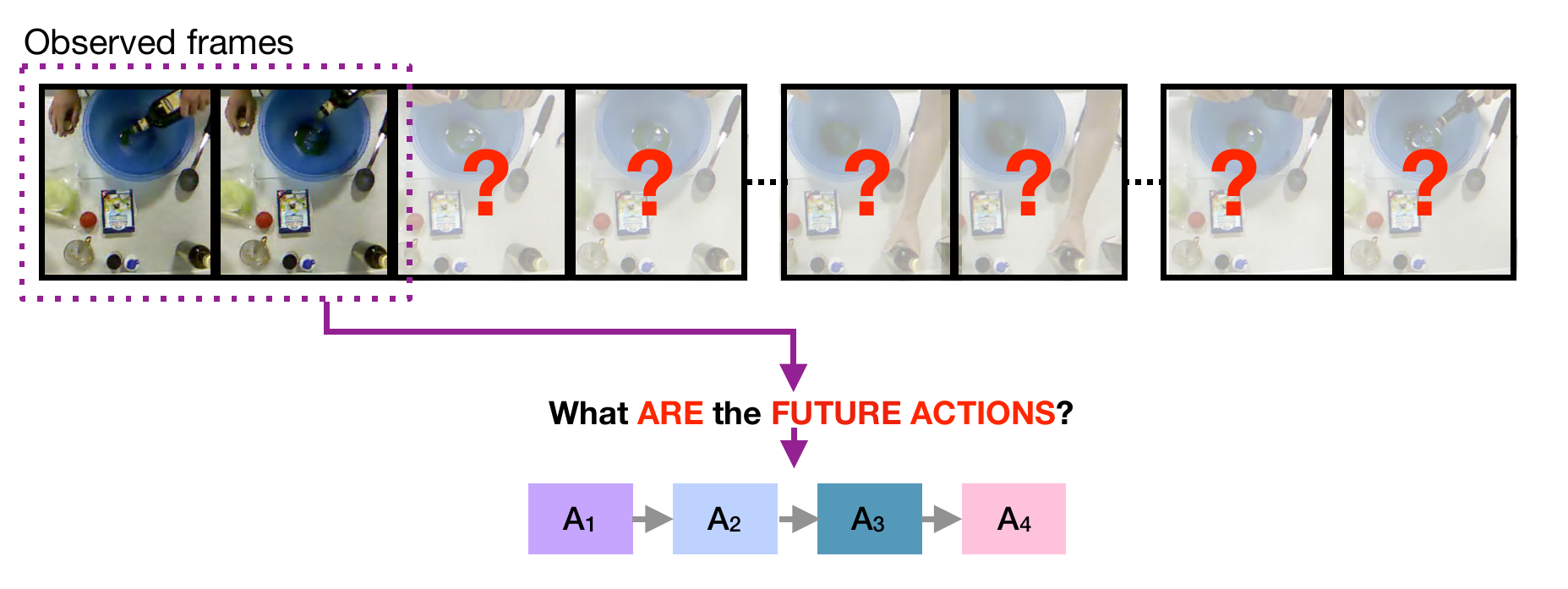}}
	\caption{Difference between action anticipation (a) and future action sequence prediction tasks. In the former task we predict what is the on going action (short-term), where as in the latter task we predict the next action sequence (long-term).}
	\label{fig:front_fig}
\end{figure}

We introduce a memory based model that predicts the next sequence of actions, by looking at only a small number of early frames. Unlike typical action anticipation methods \cite{mohammad_iccv2017} that predict the ongoing action or the next action, we aim to predict the sequence of future actions (multiple actions). Fig. \ref{fig:front_fig} illustrates the difference between the future action anticipation and future action sequence prediction tasks. In the former task, upon observing a small number of frames, we predict the ongoing action \cite{REDGao} or the next action \cite{pirri2019anticipation}. However, in future action sequence prediction we try to predict the next sequence of actions (typically for up to the next 5 minutes), after observing the first few frames. This task is more challenging as it requires us to learn long-term relationships among actions such as how some actions follow others. Despite its challenging nature, this task is highly beneficial as it can aid in predicting abnormal events and avoiding mistakes. Furthermore, it can provide aid to a human-robot interaction system to offer more efficient responses, as it is able to anticipate distant future behaviour in contrast to the action anticipation methods which can only predict at most a few seconds ahead \cite{abu2018will}. 

  Most existing and related methods utilise LSTM (Long Short-Term Memory) networks to handle video sequence information. However, as this task relies on partial information, such methods are vulnerable to ambiguities. For instance, the observed action ``wash vegetables'' could lead to numerous subsequent actions like ``cut vegetables'', ``put in fridge'', ``peel vegetables'', etc. Therefore, considering only the information from the observed input is not sufficient. It is essential to consider the current  environment context as well as the historic behaviour of the actor, and map long-term dependencies to generate more precise predictions. In our previous example, this means understanding the sequence of events preceding ``wash vegetables'' and how such event sequences have progressed in the past in order to better predict the future.

As we are dealing with longer sequences up to a duration of 5 minutes, the modelling ability of the LSTM is limited as LSTMs struggle to capture long-term dependencies when sequences are very long \cite{fernando2017tree, kumar2016ask}. To address this limitation, we make use of memory networks \cite{fernando2017tree, kumar2016ask} together with LSTM models, to improve the ability to capture long-term relationships.

Memory networks store historical facts and when presented with an input stimulus (a query) they generate an output based on knowledge that persists in the memory. The work of \cite{memoryNet} has shown encouraging results when mapping long-term dependencies among the stored facts compared to using LSTMs which map the dependencies within the input sequence. Inspired by these findings we incorporate neural memory networks and propose a framework for generating long-term predictions for the action sequence prediction task. In addition, we model these dependencies within different input streams separately through individual memory models.

\begin{figure}[htbp]
        \centering
        	\includegraphics[width=0.53\linewidth]{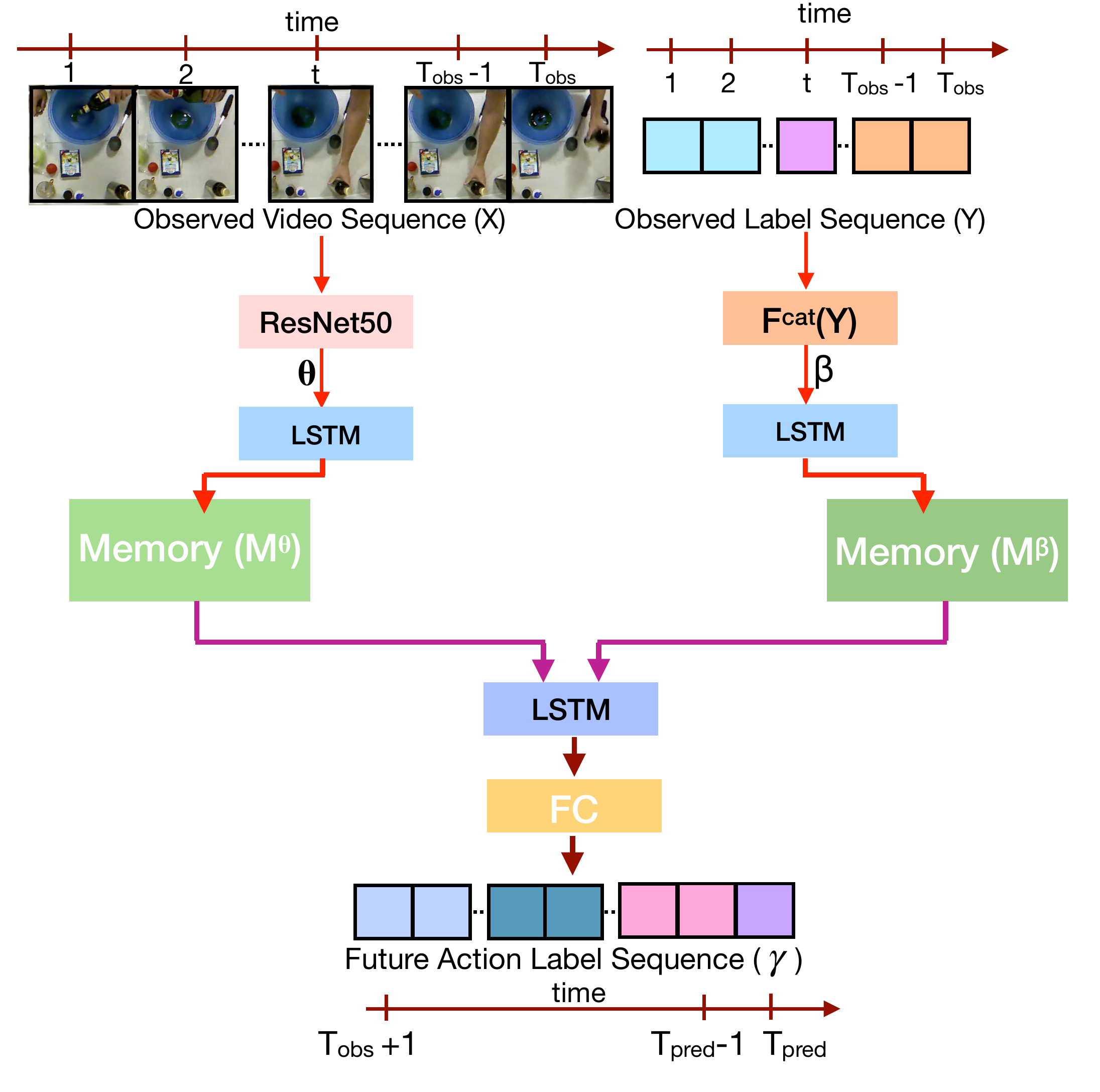}
	\vspace{-3mm}
	\caption{Proposed action sequence forecasting model: We use the observed frame sequence (X) and the observed label sequence (Y) as our inputs. These inputs are first encoded through ResNet50 and a categorical function $F^{cat}$ and is then temporally mapped using LSTMs. We use two memory modules to model the long-term relationships of the individual modalities, and they output relevant facts given an input query. These individual memory outputs are concatenated and passed through another LSTM and a fully-connected layer, generating the future action sequence. The functionality of $M^{\theta}$ and $M^{\beta}$ is described in Sec. \ref{sec:memory_networks}}
	\label{fig:arc}
\end{figure}

Fig. \ref{fig:arc} shows the overall architecture of our proposed model. The model is fed with two input streams: the observed frame sequence ($X$) and, following \cite{abu2018will},  the corresponding labels of the observed frames ($Y$). The observed frames are passed through a ResNet50 network \cite{resnet} pre-trained on ImageNet \cite{imageNet} and the extracted ($\theta$) features are passed through a separate LSTM layer. The output of the LSTM layer is then passed to the memory module, $M^{\theta}$. The observed frame-label sequence is converted to a categorical representation and passed through a seperate LSTM and then a second memory module, $M^{\beta}$. The output sequences from the two memory networks are merged and passed through a third LSTM layer followed by a fully-connected layer for the final anticipation task. As such, the network can learn how to extract long-term information from each mode separately, and can learn how best to combine this information.

The main contributions of this work can be summarised as follows:
\begin{itemize}
\item We introduce a novel neural memory network based framework for the prediction of a future action sequence.
\item We demonstrate the utility of using both visual and action label features for this task.
\item We effectively model long-term dependencies of individual streams through seperate memory modules, enabling better fusion of the salient spatio-temporal features. 
\item We perform extensive evaluations on multiple public benchmarks and show that our proposed method outperforms the state-of-the-art methods by a significant margin.
\end{itemize}

\vspace{-3mm}
\section{Related Work}
\vspace{-3mm}
Action recognition \cite{twoStream, gammulle2018multi, gammulle2017} and segmentation \cite{lea2016seg, gammulle2019coupled} are popular topics in computer vision, where the operations are performed over fully observed sequences after the occurrence of the actual action or actions. As such, these methods are limited to post-event analysis applications. In contrast, methods that provide early action predictions have much greater utility as they can allow a system to respond in advance to an action as it occurs, and thus can be used in interactive situations (i.e. human robot interaction), or to potentially detect and mitigate risks.

\textbf{Future Action Prediction: }These methods attempt predict the ongoing action as early as possible from a limited portion of the video frames. Aliakbarian et al. \cite{mohammad_iccv2017} introduced a multi-layer LSTM model that incorporates a loss function in order to improve the anticipation task. Similarly, \cite{ma2016learning} introduced a novel ranking loss that is used together with the classification loss to train the proposed LSTM model. In \cite{mahmud2017joint}, the introduced deep network is fed with both visual and previous activity features that are passed through an LSTM to map the temporal relations and predict the next action label together with the starting time. However, these methods anticipate actions only for the next few seconds. The work presented in \cite{abu2018will} introduces a novel paradigm for future action sequence prediction and proposed a method to anticipate activities that occur within a time horizon of up to 5 minutes, with the aid of two models which are based on RNNs and CNNs respectively. Such long-term prediction of activities is important as these can predict multiple future actions, and there is strong potential for such systems to be used in real world tasks such as providing warning systems for security, or within aged care facilities. 

However, none of the existing works in either action anticipation or future action sequence prediction have investigated how best to capture long-term dependencies. We speculate that LSTMs fail to capture such dependencies as they model only the relationships within a given sequence \cite{fernando2017tree}. To address this, we demonstrate a method to effectively capture these relationships using neural memory networks. 

\textbf{Neural Memory Networks: } When anticipating future actions it is essential to be equiped with a model that can map long-term relationships among the observed actions. Memory networks can facilitate this process and these have been widely used in different areas of computer vision \cite{memoryNet, fernando2017tree} to model and respond to such long-term temporal relationships. However, in the area of action recognition only a limited number of methods are supported by memory networks \cite{xie2018memory, pirri2019anticipation}. 

In \cite{xie2018memory} the authors demonstrate the utility of storing temporal information in a memory cell for the task of human action recognition. However, in this work they have only considered the temporal relationships within the given input sequence. The authors of \cite{pirri2019anticipation} have investigated the effect of NMNs for future action anticipation where they store the internal state of the LSTM cell. They demonstrate that stored internal states in the memory contribute favourably when mapping long-term relationships for the action anticipation task. 

However, none of these works have investigated the application of external memories for storing the long-term relationships between sequences. These relationships are also of importance as it allows the model to oversee the current context as well as the temporal evolution of the environment, enabling better anticipation of future actions. We demonstrate how multiple memory modules can be introduced into the network architecture to effectively capture the dependencies between different input modalities and perform better fusion of multiple feature modalities. This allows the learning framework to understand how prominent each modality is in the current environmental context and to effectively utilise multiple modalities for the action anticipation task.  

Furthermore, in both \cite{xie2018memory} and \cite{pirri2019anticipation} the authors have relied on human skeletal data which is not readily available in most real world application settings. In contrast, we utilise only a portion of the RGB input video and the labelled action classes for this portion, and predict the future action sequence. To the best of our knowledge this is the first work that introduces external neural memory networks for future action anticipation.          
 
\vspace{-3mm}
\section{Methodology}
\vspace{-3mm}
We address the problem of future action sequence prediction, using both the observed frames and corresponding action labels to predict the future behaviour.

Our problem can be mathematically formulated as follows. Let the set of observed frames of the video be $X = {x_1, x_2,\cdots, x_{\textsc{T}_{obs}}}$, where $T_{obs}$ is the observed frame count. The second input is the corresponding labels of the observed frame sequence, $Y = {y_1,y_2, \cdots, y_{\textsc{T}_{obs}}}$. 

Prior to feeding the inputs to the network we extract CNN feature embeddings for $X$,
\begin{equation}
\theta = f^{ResNet}(X),
\label{eq:1}
\end{equation} 

and the input, $Y$, is converted into categorical form (i.e. a sequence of one-hot vectors),
\begin{equation}
\beta = f^{cat}(Y).
\label{eq:2}
\end{equation} 
\vspace{-3mm}
\textbf{Problem definition: } Given $\theta$ and $\beta$, predict the future action sequence $\gamma$,  
\begin{equation}
f([\theta, \beta]) = \gamma.
\label{eq:3}
\end{equation} 

Here, defining the predicted frame count as $T_{pred}$, $\gamma$ can be further defined as,
\begin{equation}
\gamma = f^{cat}({y_{\textsc{T}_{obs}+1}},\cdots, y_{\textsc{T}_{pred}}).
\label{eq:4}
\end{equation}

To achieve this task, our proposed method encodes the input features, $\theta$ and $\beta$ using separate LSTM layers, such that, 
\begin{equation}
h^{\theta}_t = f^{LSTM}(\theta_t), \hspace{2mm} h^{\beta}_t = f^{LSTM}(\beta_t),
\label{eq:5}
\end{equation} 
and uses memory networks to extract salient information from each stream.
\vspace{-3mm}
\subsection{Memory Networks}
\label{sec:memory_networks}
In our proposed architecture we utilise two memory networks, namely $M^{\theta}$ and $M^{\beta}$ for the individual input streams: the observed feature sequence ($\theta$) and the corresponding observed labels ($\beta$). $M^{\theta}_{t-1}$ and $M^{\beta}_{t-1}$ are the states of the memories, $M^{\theta}$ and $M^{\beta}$, at time instance $t-1$ respectively.

\begin{figure}[!h]
        \centering
        	\includegraphics[width=0.45\linewidth]{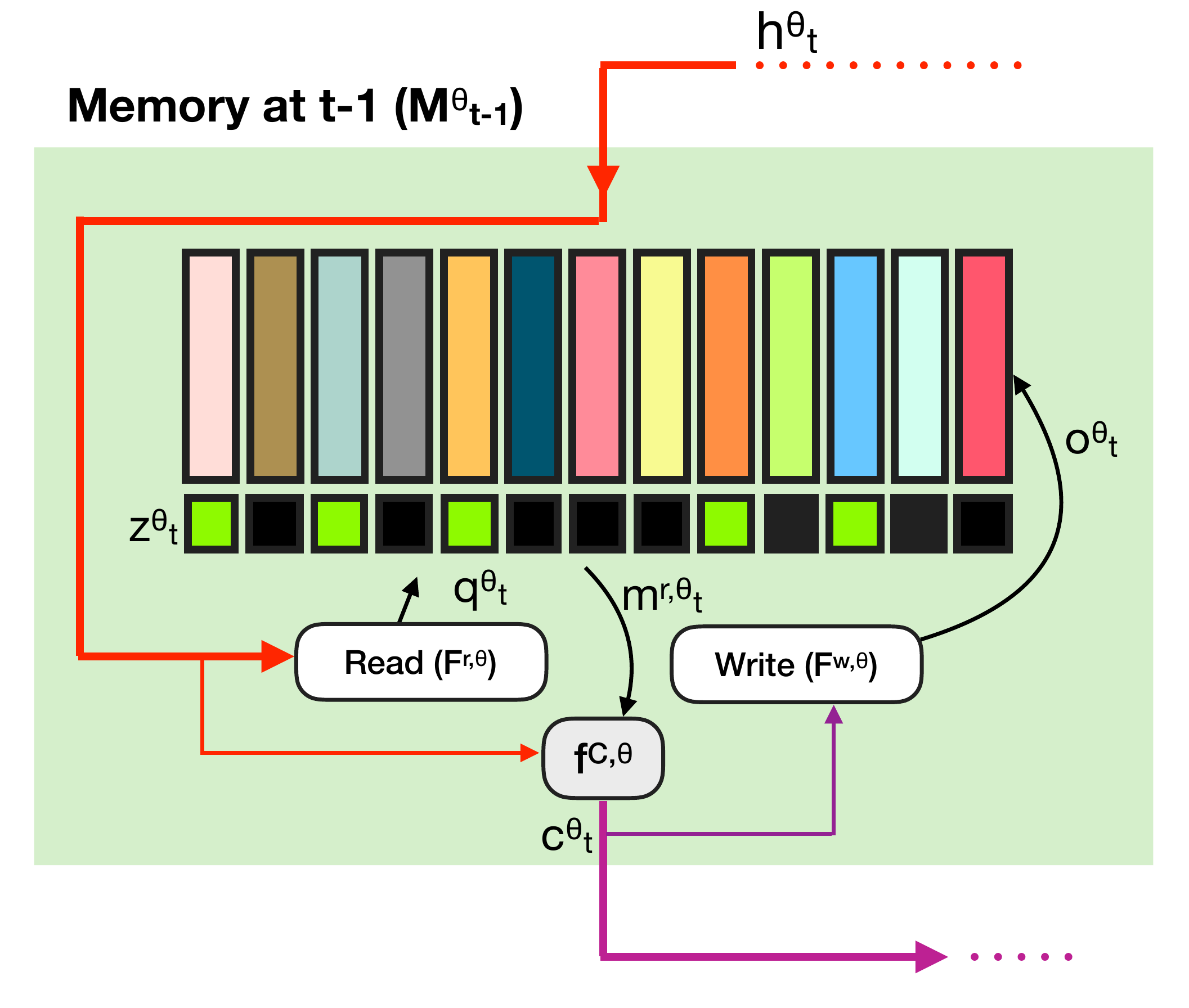}
	\vspace{-3mm}
	\caption{Memory Network: The state of the memory at time instance t-1 is $M^{\theta}_{t-1}$. The read function, $f^{r, \theta}$, receives the encoded hidden state, $h^{\theta}_t$, of the LSTM at time instance $t$ and generates a vector, $q_t^{\theta}$, to query the memory. We generate an attention score vector, $z^{\theta}_t$, quantifying the similarity between $q_t^{\theta}$ and the content of each slot of $M^{\theta}_{t-1}$ and generate the output of the read function, $m^{r, \theta}_t$. This is subsequently passed through a MLP, generating the output $c^{\theta}_t$. Finally, the write function, $f^{w, \theta}$, updates the memory and propagates it to the next time step.  }
	\label{fig:memory}
\end{figure}

Each memory can be defined as, $M \in \mathbb{R}^{l\times{k}}$, where there are $l$ slots, each of which contains an embedding of length $k$. When utilising memory, there are two main operations: the read operation and the write operation. Fig. \ref{fig:memory} illustrates these two operations which will be discussed in the following sub-sections.

\textbf{Memory Read Operation: } 
Given the encoded hidden state, $h^{\theta}_t $, from the LSTM encoder in Eq. \ref{eq:5}, the read operation $f^{r,\theta}$ generates a query $q^{\theta}_t$, to question $M^{\theta}_{t-1}$ such that,
\begin{equation}
\vspace{-0.5mm}
q^{\theta}_t = f^{r,\theta}(h^{\theta}_t).
\label{eq:7}
\end{equation} 

Motivated by \cite{memoryNet}, we implement $f^{r,\theta}$ using an LSTM cell. We attend to each memory slot in $M^{\theta}_{t-1}$ and using a softmax function we quantify the similarity between the content stored in each slot and the query vector $q^{\theta}_t$ such that,
       
\begin{equation}
\vspace{-1mm}
z^{\theta}_t = softmax([q^{\theta}_t]^{T} M^{\theta}_{t-1}).
\label{eq:8}
\end{equation} 

We multiply the content of the memory slots with the respective score values, $z^{\theta}_{t}$, and generate a vector $m^{r, \theta}_t$, 
\begin{equation}
\vspace{-1mm}
m^{r, \theta}_t = z^{\theta}_t M^{\theta}_{t-1},
\label{eq:9}
\end{equation} 
 
which is subsequently passed through a multi-layer perceptron \cite{pal1992multilayer} to generate the memory output, 
\begin{equation}
\vspace{-1mm}
c^{\theta}_t = f^{c, \theta}(h^{\theta}_t, m^{r,\theta}_t).
\label{eq:10}
\end{equation} 

This function determines what portion from the information of the current memory input $h^{\theta}_t$ and the historical information stored in the memory should be output at the current time instant by the read function. Ideally, we wish for $c^{\theta}_{t}$ to capture salient information from both the input and stored history that can be used to predict the future behaviour.

\textbf{Memory Update Operation: } 
Motivated by the update procedure of \cite{memoryNet}, first we pass the memory output, $c^{\theta}_t$, through a write function, $f^{w, \theta}$, to generate a vector to update the memory,

\begin{equation}
\vspace{-1mm}
o^{\theta}_t = f^{w, \theta}(c^{\theta}_t).
\label{eq:11}
\end{equation} 
Similar to $f^{r, \theta}$, we implement $f^{w, \theta}$ using an LSTM cell. Using this vector and the score vector $z^{\theta}_t$ derived from Eq. \ref{eq:8}, we update the content of each memory slot based on the informativeness reflected in the score vector such that,
\begin{equation}
\vspace{-1mm}
M^{\theta}_t = M^{\theta}_{t-1}(I - (z^{\theta}_t \otimes{e_{k^{\theta}}})^T) + (o^{\theta}_t \otimes{e_{l^{\theta}}})(z^{\theta}_t \otimes{e_{k^{\theta}}})^T, 
\label{eq:12}
\end{equation} 
where $I$ is a matrix of ones, $e_{l^{\theta}} \in \mathbb{R}^{l^{\theta}}$ and $e_{k^{\theta}} \in \mathbb{R}^{k^{\theta}}$ are vectors of ones and $\otimes$ denotes the outer vector product which duplicates its left vector $l^{\theta}$ or $k^{\theta}$ times, $l^{\theta}$ is the number of memory slots in $M^{\theta}$, and $k^{\theta}$ is the embedding dimension of each slot in $M^{\theta}$. 
Similarly, for $M^{\beta}_{t-1}$, we define the read and write operations such that,
\begin{equation}
q^{\beta}_t = f^{r,\beta}(h^{\beta}_t),
\label{eq:13}
\end{equation}
\vspace{-2mm}     
\begin{equation}
z^{\beta}_t = softmax([q^{\beta}_t]^{T} M^{\beta}_{t-1}),
\label{eq:14}
\end{equation} 
\vspace{-2mm}   
\begin{equation}
m^{r, \beta}_t = z^{\beta}_t M^{\beta}_{t-1},
\label{eq:15}
\end{equation} 
\vspace{-2mm}
\begin{equation}
c^{\beta}_t = f^{c, \beta}(h^{\beta}_t, m^{r,\beta}_t),
\label{eq:16}
\end{equation} 
\vspace{-2mm} 
\begin{equation}
o^{\beta}_t = f^{w, \beta}(c^{\beta}_t),
\label{eq:17}
\end{equation} 
\vspace{-2mm}
\begin{equation}
M^{\beta}_t = M^{\beta}_{t-1}(I - (z^{\beta}_t \otimes{e_{k^{\beta}}})^T) + (o^{\beta}_t \otimes{e_{l^{\beta}}})(z^{\beta}_t \otimes{e_{k^{\beta}}})^T, 
\label{eq:18}
\end{equation} 
where $h^{\beta}_t$ is the encoded hidden state from the LSTM encoder function in Eq. \ref{eq:5}.

\subsection{Forecasting Future Action Sequence}
\vspace{-3mm}   
Concatenating the outputs $c^{\theta}_t$ and $c^{\beta}_t$ of the memory read operations for time instance $t$ we generate an augmented vector, 
\begin{equation}
\vspace{-1mm}
S_t = [c^{\theta}_t, c^{\beta}_t],
\label{eq:19}
\end{equation} 
which is subsequently passed through an LSTM function,
\begin{equation}
\vspace{-1mm}
h'_t = f^{LSTM}(S_t).
\label{eq:20}
\end{equation} 
 The final classification is obtained by passing this hidden vector $h'_t$ through a fully-connected layer with softmax activation,
\begin{equation}
\vspace{-1mm}
\gamma_t = f^{FC}(h'_t).
\label{eq:21}
\end{equation} 
It should be noted that the above process is iteratively applied, feeding the previous time-step's prediction back to the memory, to generate a sequence of future action predictions.

\vspace{-3mm}
\section{Experiments}
\label{sec:exp}
\vspace{-3mm}   
This section includes details regarding the implementations, datasets, evaluation results and the discussion. Due to the page limitation, hyper-parameter evaluation, time complexity and qualitative results are included in the supplementary materials.   

\subsection{Implementation Details}
The first input, the observed frame sequence ($X$) is passed through the ResNet50 \cite{resnet} network which is pre-trained on ImageNet \cite{imageNet}. The features are extracted from the \textit{activation\_50} layer of the ResNet50 network and these features are then passed through a LSTM layer with a hidden state dimension of 300. The LSTM layer output sequence is passed through a memory network ($M^{\theta}$) with the memory length, $l^{\theta} = 24 $ and with the feature dimensionality of $k^{\theta} = 300$. 

Similarly, the second input stream ($Y$) is passed through a separate LSTM layer with a hidden dimension of 30. Then these outputs are passed through a seperate memory, $M^{\beta}$, where the length of the memory $l^{\beta} = 20$, and the feature dimensionality $k^{\beta}= 30$. The memory outputs, $c^{\theta}$ and $c^{\beta}$ are concatenated together and passed through a third LSTM layer with a hidden state dimensionality of 300. 

\vspace{-3mm}
\subsection{Datasets}
\label{sec:datasets}

To achieve a fair comparison with the baseline approach in \cite{abu2018will}, we utilise the same datasets in our evaluation: the Breakfast \cite{breakfast} and the 50 Salads \cite{50salads} datasets. These datasets have been widely used for fine-grained action segmentation \cite{lea2016seg} which are based on fully observed video sequences.

\textbf{Breakfast dataset \cite{breakfast}} is composed of 1712 videos containing 52 subjects performing breakfast preparation activities. The videos are recorded in 18 different kitchens and are composed of 48 fine-grained actions. Similar to \cite{abu2018will}, we also utilise the four splits provided. 

\textbf{50 Salads dataset \cite{50salads}} contains 50 videos of salad preparation activities performed by 25 actors where each actor prepares two salads. The dataset is composed of 17 fine-grained action classes. For the evaluation, we utilise a five-fold cross validation.

\vspace{-3mm}
\subsection{Results}
\label{sec:results}
We follow the experimental approach of \cite{abu2018will}. To the best of our knowledge, the method of \cite{abu2018will} is the first and the only method that predicts the sequence of future actions. We perform comparisons to their introduced RNN and CNN models. Additionally, we use the grammar \cite{richard2017weakly} and Nearest Neighbour Search (NNS) methods which are also reported in \cite{abu2018will}. Results on Breakfast and 50 Salads are shown in Tab. \ref{tab:tab_1} and Tab. \ref{tab:tab_2} respectively. Similar to \cite{abu2018will}, we report accuracies when observing different percentages of input frames (Observed \%) and predicting different lengths into the future (Predicted \%) from that point onwards. 

\begin{table}[htbp]
\centering
\resizebox{\linewidth}{!}{
\begin{tabular}{|c|c|c|c|c|c|c|}
\hline
\multicolumn{1}{|l|}{Observed \%} & Predicted \% & Proposed & CNN \cite{abu2018will}    & RNN \cite{abu2018will}    & NNS \cite{abu2018will}     & Grammer \cite{richard2017weakly} \\ \hline
\multirow{4}{*}{20\%}             & 10\%         &  \textbf{87.20}  & 57.97 & 60.35 & 43.78 & 48.92  \\ \cline{2-7} 
                                  & 20\%         & \textbf{85.24} & 49.12 & 50.44 & 37.26 & 40.33  \\ \cline{2-7} 
                                  & 30\%         &  \textbf{81.02}  & 44.03 & 45.28 & 34.92 & 36.24  \\ \cline{2-7} 
                                  & 50\%         &  \textbf{75.47} & 39.26 & 40.42 & 29.84 & 31.46  \\ \hline
\multirow{4}{*}{30\%}             & 10\%         & \textbf{87.90} & 60.32 & 61.45 & 44.12 & 52.66  \\ \cline{2-7} 
                                  & 20\%         &\textbf{85.79}& 50.14 & 50.25 & 37.69 & 42.15  \\ \cline{2-7} 
                                  & 30\%         &\textbf{82.10}& 45.18 & 44.90 & 35.70 & 38.44  \\ \cline{2-7} 
                                  & 50\%         & \textbf{76.30}    & 40.51 & 41.75 & 30.19 & 33.09  \\ \hline
\end{tabular}}

\vspace{1mm}
\caption{The evaluation results of the proposed model on the Breakfast dataset \cite{breakfast}.}
\label{tab:tab_1}
\end{table}
\begin{table}[htbp]
\centering
\resizebox{\linewidth}{!}{
\begin{tabular}{|c|c|c|c|c|c|c|}
\hline
\multicolumn{1}{|l|}{Observed \%} & Predicted \% & Proposed & CNN \cite{abu2018will}    & RNN \cite{abu2018will}   & NNS \cite{abu2018will}    & Grammer \cite{richard2017weakly} \\ \hline
\multirow{4}{*}{20\%}             & 10\%         &   \textbf{69.97}       & 36.08 & 42.30 & 25.21 & 28.69  \\ \cline{2-7} 
                                  & 20\%         & \textbf{64.33} & 27.62 & 31.19 & 21.05 & 21.65  \\ \cline{2-7} 
                                  & 30\%         & \textbf{62.71}& 21.43 & 25.22 & 16.34 & 18.32  \\ \cline{2-7} 
                                  & 50\%         & \textbf{52.16}& 15.48 & 16.82 & 13.17 & 10.37  \\ \hline
\multirow{4}{*}{30\%}             & 10\%         &\textbf{68.10}& 37.36 & 44.19 & 22.12 & 26.71  \\ \cline{2-7} 
                                  & 20\%         & \textbf{62.29}& 24.78 & 29.51 & 17.15 & 14.59  \\ \cline{2-7} 
                                  & 30\%         & \textbf{61.18} & 20.78 & 19.96 & 18.38 & 11.69  \\ \cline{2-7} 
                                  & 50\%         & \textbf{56.67}& 14.05 & 10.38 & 14.71 & 09.25  \\ \hline
\end{tabular}}

\vspace{1mm}
\caption{The evaluation results of the proposed model on the 50 Salads dataset \cite{50salads}.}
\label{tab:tab_2}
\end{table}

Similar to \cite{abu2018will}, in the experiments as the input Y (Eq. 2) we use the ground truth observed labels provided with the dataset. However, in Sec \ref{sec:sens}, we conduct additional experiments using the class labels generated using the method of \cite{gammulle2019coupled}.
   
We speculate that the significant improvement in the results for our proposed model compared to the CNN and RNN models in \cite{abu2018will} is mainly due to the long-term dependency modelling enabled by the utilisation of the external memory networks in our approach. The RNN model in \cite{abu2018will} predicts the next action class and to handle the long-term prediction task the predicted features are fed back to the network. Even though the RNN model has the ability capture temporal information, still it considers only the relationships within the current sequence due to the internal memory structure, making accurate long-term prediction intractable. Similarly the CNN based model of \cite{abu2018will}, which first maps the observed examples to a matrix representation and trains the model to predict the sequence of future actions without any temporal modelling, performs worse compared to the RNN method reported in \cite{abu2018will}. In contrast, the memory network proposed in our work is capable of capturing both short-term within sequence dependencies as well as long-term between sequence relationships when making a prediction. This allows the model to learn and store overall patterns of behaviour in the memory.

We further compare the performance for different observed/ predicted sequence lengths. We observe a significant performance degradation for both the CNN and RNN baseline models in \cite{abu2018will}, when predicting lengthier future sequences. For instance in Tab. \ref{tab:tab_2} we observe that the performance of the RNN based method reported in \cite{abu2018will} drops from 44.19\% to 10.38\% when the length of the predicted sequence increases from 10\%  to 50\% whereas for our approach the corresponding drop is from 68.10\% to 56.67\%, which is much smaller. This clearly demonstrates that the relationships captured through the RNN are insufficient to perform accurate forecasting. On the other hand, in the proposed architecture by capturing both short-term and long-term dependencies we attain better modelling of the current context of the environment and how it evolves over time, and obtain much better performance when anticipating actions in the distance future. Even though we observe a slight degradation in performance when predicting 50\% of the future actions, the performance degradation is considerably less severe compared to \cite{abu2018will}.


To further illustrate the our method we evaluated a series of ablation models as follows:
\begin{itemize}
\item a) $\theta$: Uses only the $\theta$ input stream and uses only the encoding and decoding LSTMs followed by a fully-connected layer.
\vspace{-1mm}
\item b) $\beta$: Similar to (a) but uses the $\beta$ input stream.
\vspace{-1mm}
\item c) $\theta + M^{\theta}$ Similar to (a) but uses a neural memory to store long-term relationships.
\vspace{-1mm}
\item d) $\beta + M^{\beta}$ Similar to (c) but uses the $\beta$ input stream.
\vspace{-1mm}
\item e) $\theta + \beta + M$ Similar to the proposed method but uses only one memory component (i.e. $\theta$ and $\beta$ features are concatenated prior to being fed to the memory). 
\end{itemize}

The ablation results on Breakfast and 50 Salads datasets are presented in Tab. \ref{tab:tab_4} and \ref{tab:tab_3}, respectively. In these we observe 30\% of the video and predict 50\% of the future.

\begin{table}
\parbox{.45\linewidth}{
\centering
\begin{tabular}{|c|c|}
\hline
Model        & Accuracy \\ \hline
a) $\theta$ & 21.56\\\hline
b) $\beta$ &  40.45 \\\hline
c) $\theta + M^{\theta}$& 51.29     \\ \hline
d) $\beta + M^{\beta}$ & 68.09    \\ \hline
e)  $\theta + \beta + M$  & 72.00    \\ \hline
Proposed     & \textbf{76.30}    \\ \hline
\end{tabular}
\caption{Ablation evaluation results on the Breakfast dataset.}
\label{tab:tab_4}
}
\hfill
\parbox{.45\linewidth}{
\centering
\begin{tabular}{|c|c|}
\hline
Model        & Accuracy \\ \hline
a) $\theta$ & 13.76\\\hline
b) $\beta$ &  21.33 \\\hline
c) $\theta + M^{\theta}$&   37.44  \\ \hline
d) $\beta + M^{\beta}$  &  48.21   \\ \hline
e)  $\theta + \beta + M$  & 50.73    \\ \hline
Proposed     & \textbf{56.67}    \\ \hline
\end{tabular}
\caption{Ablation evaluation results on the 50 Salads dataset.}
\label{tab:tab_3}
}
\end{table}

Similar to \cite{abu2018will},we observe that the input label sequence is the prominent stream. This is because having only the observed frame sequence, the network first needs to understand what the current actions are before predicting the next action sequence. With the observed labels as inputs, this task becomes much easier as it can effectively skip the current action recognition step. Almost every daily activity is composed of a related set of sub-actions following one after another, and there are always some actions that are more likely to appear next. Hence, knowing the labels of the observed actions makes the final task less complex. 

With the introduction of the memory component (i.e models c and d) we observe improved performance as it provides more capacity for the model to map long-term relationships. Furthermore, with the fusion of the two input modalities ( i.e model e) we are able to capture complimentary information of the individual modalities. However, this is still suboptimal as two separate modes, each of which encodes information in a different way, are combined within a single memory. With the introduction of two seperate memories, the proposed model is able to better model each stream, and fuse the streams while considering how each is evolving over time. We note that while the action class is the dominant stream, the fused system achieves a substantial performance improvement suggesting that action labels alone are not enough to predict future behaviour. We hypothesise that the visual stream is able to extract additional scene cues that provide complementary context information to aid prediction.

\subsection{Sensitivity Analysis}
 \label{sec:sens}
 In order to estimate the sensitivity of the proposed method on the preciseness of the input action labels we conducted an analysis using the action labels which are predicted using the action segmentation model of \cite{gammulle2019coupled}. We pass the observed potion of the video through the segmentation model of \cite{gammulle2019coupled} and generate the corresponding action labels for those frames. These observed frames and the generated labels are then fed to the proposed method. We conducted this experiment using the test set of the 50 Salads dataset \cite{breakfast}. 
 
 The evaluation results are presented in Tab. \ref{tab:sensitivity}. Similar to the ablation evaluations we observe 30\% of the video and predicted 50\% of the future. Even though we observe a slight degradation of the performance when using the labels generated from the method of \cite{gammulle2019coupled}, instead of using the ground truth action labels, still the performance is significantly superior compared to all the baselines which use ground truth action labels.
 
\begin{table}[htbp]
\centering
\begin{tabular}{|c|c|}
\hline
Model        & Accuracy \\ \hline
CNN \cite{abu2018will}  with ground truth labels &  14.05 \\ \hline 
RNN \cite{abu2018will} with ground truth labels &  10.38 \\ \hline 
NNS \cite{abu2018will} with ground truth labels &  14.71 \\ \hline \hline
Proposed method with labels of \cite{gammulle2019coupled} & 49.21 \\ \hline
Proposed method with ground truth lables   & \textbf{56.67}    \\ \hline
\end{tabular}
\caption{Sensitivity analysis on the 50 Salads dataset \cite{breakfast} using the action labels predicted using the method of \cite{gammulle2019coupled}.}
\label{tab:sensitivity}
\end{table}

\vspace{-3mm}
\section{Conclusion}
\vspace{-3mm}
We have introduced a neural memory network based model for forecasting the next action sequence in fine-grained action videos. The proposed system eliminates the deficiencies in current state-of-the-art RNN based temporal modelling which only considers the within sequence relationships. The proposed system is able to map the long-term dependencies in the entire data domain. Furthermore, by utilising individual memories for the two inputs, the observed frames and the corresponding action labels, we are able to capture different information cues to support the prediction task. Through our extensive experimental evaluations we demonstrate the utility of this fusion strategy, where we outperform by a significant margin the current state-of-the-art techniques on multiple public benchmarks.

\bibliography{paper7}
\end{document}